\documentclass{article}

\usepackage[T1]{fontenc}
\usepackage{lmodern}

\usepackage{microtype}
\usepackage[pdftex]{graphicx}
\usepackage{subcaption}
\usepackage{caption}
\usepackage{booktabs} 
\usepackage{amssymb}
\usepackage{amsmath}

\usepackage{hyperref}

\usepackage[utf8]{inputenc} 
\usepackage[T1]{fontenc}    
\usepackage{hyperref}       
\usepackage{url}            
\usepackage{booktabs}       
\usepackage{amsfonts}       
\usepackage{nicefrac}       
\usepackage{microtype}      

\usepackage{authblk}

\title{Learning Neuron Non-Linearities with \\ Kernel-Based Deep Neural Networks}

\DeclareMathOperator{\pa}{pa}

\def\(#1){[\hbox{$\mkern1mu\thickmuskip=\thinmuskip#1\mkern1mu$}]} 

\author[1,2]{Giuseppe Marra}
\author[1,2]{Dario Zanca }
\author[1,2]{\\Alessandro Betti}
\author[2]{ Marco Gori}
\affil[1]{DIISM, University of Siena}
\affil[2]{DINFO, University of Florence}

\begin{document}

\maketitle

\begin{abstract}
  	The effectiveness of deep neural architectures has been widely supported
  	in terms of both experimental and foundational principles. 
  	There is also clear evidence that the activation function 
  	(e.g. the rectifier and the LSTM units)  plays a crucial role in 
  	the complexity of learning.
  	Based on this remark, this paper discusses an optimal selection of 
  	the neuron non-linearity in a functional framework that is inspired from
  	classic regularization arguments. It is  shown that the best  activation function 
  	is represented by a kernel expansion in the training set, that can 
  	be effectively approximated over an opportune set of points 
  	modeling 1-D clusters. The idea can be naturally extended to 
  	recurrent networks, where the expressiveness of kernel-based activation
  	functions turns out to be a crucial ingredient to capture long-term 
  	dependencies. We give experimental evidence of this property by a set 
  	of challenging experiments, where we  compare the  results with 
  	neural architectures based on state of the art LSTM cells. 
\end{abstract}

\section{Introduction}
\label{introduction}
By and large, the appropriate selection of the activation function in deep 
architectures is regarded as an important choice for achieving challenging 
performance. For example, the rectifier function~\cite{journals/jmlr/GlorotBB11} has been playing an important role 
in the impressive scaling up of nowadays deep nets. Likewise, 
LSTM cells~\cite{Hochreiter:1997:LSM} are widely recognized as the most important ingredient to face
long-term dependencies when learning by recurrent neural networks. 
Both choices  come from insightful ideas on the actual non-linear process
taking place in deep nets. 
At a first glance,  one might wonder why such an optimal choice must be 
restricted to a single unit instead of extending it to the overall
function to be 
learned. In addition, this general problem has been already been solved;
its solution~\cite{Poggio_ieee90,PoggioGirosiNC95,EPP1999} 
is in fact at the basis of kernel machines, 
whose limitations as shallow nets, 
have been widely addressed (see e.g.~\cite{DBLP:journals/nature/LeCunBH15,DBLP:journals/corr/MhaskarLP16}). 
However, the  optimal formulation given for the neuron non-linearity 
enjoys the tremendous advantage of acting on 1-D spaces.
This strongly motivates the reformulation of the problem of learning in 
deep neural network as a one where the weights and the activation functions
are jointly determined by optimization in the framework of regularization
operators~\cite{SS-NN-1998}, that are used to enforce the smoothness of the solution. 
The idea of learning the activation function is not entirely new. In \cite{turner2014neuroevolution}, activation functions are chosen from a pre-defined set and combine this strategy with a single scaling parameter that is learned during training. It has been argued that
one can think of this function  as a neural network itself, so as 
the overall architecture is still characterized by a directed
acyclic graph~\cite{Castelli:2014:CSU}. Other approaches learn activation functions as piecewise linear  \cite{agostinelli2014learning}, doubled truncated gaussian \cite{NIPS2017_7035} or Furier series \cite{eisenach2016nonparametrically}.
In this paper, it is proven that, like for kernel
machines, the optimal solution can be expressed by a kernel expansion, 
so as the overall optimization is reduced to the discovery of a finite set
of parameters. The risk function to be minimized contains
the weights of the network connections, as well as the parameters
associated with the  the points of the kernel expansion. Hence, 
the classic learning of the weights of the network takes place with the
concurrent development of the optimal shape of the activation functions, 
one for each neuron.  As a consequence, the machine architecture turns out to enjoy 
the strong representational issues of deep networks in high dimensional
spaces that is conjugated with the elegant and effective
setting of kernel machines for the learning of the activation functions. 
The powerful unified regularization framework is not the only feature
that emerges from the proposed architecture. Interestingly, 
unlike most of the activation functions used in deep networks,
those that are typically developed during learning, are not necessarily monotonic. 
This property has a crucial impact in their adoption in classic recurrent networks,
since this properly addresses classic issues of gradient vanishing
when capturing long-term dependencies. 
Throughout this paper, recurrent networks 
with activation functions based on kernel expansion, are referred to 
as Kernel-Based Recurrent Networks (KBRN).
The intuition is that the associated iterated map can either be contractive or expansive.
Hence, while in some states the contraction yields gradient vanishing,
in others the expansion results in to gradient pumping, which allows the
neural network to propagate information back also in case
of long time dependences.  
The possibility of implementing contractive
and expanding maps during the processing of a given sequence comes
from the capabilities of KBRN to develop different activation functions for different
neurons that are not necessarily monotonic. This variety of units is 
somewhat related to the clever solution proposed in LSTM cells~\cite{Hochreiter:1997:LSM},
where the authors early realized  that there was room for getting rid of the 
inherent limitation of the contractive maps deriving from sigmoidal units.
The given experimental results provide
evidence of this property on challenging benchmarks that are inspired
to seminal paper~\cite{Bengio_trnn93}, where the distinctive information
for classification of long sequences is only located in the first positions,
while the rest contains uniformly distributed noisy information. 
We get very promising results on these benchmarks when comparing KBRN 
with state of the art recurrent architectures based on LSTM cells.

\section{Representation and learning}
\label{theory}
The feedforward architecture that we consider is based on
a  directed graph $D \sim (V,A)$, where $V$ is the set of ordered 
vertices and $A$ is the set of the oriented arcs. Given $i, j \in V$
there is connection from $i$ to $j$ iff $i \prec j$. Instead of assuming
a uniform activation function for each vertex of $D$, a specific 
function $f$ is attached to each vertex.
We  denote with $I$ the set
of input neurons, with $O$ the set of the output neurons and
with $H=V\setminus(I\cup O)$ the set of hidden neurons;
the cardinality of these sets will be denoted as $|I|$, $|O|$,
$|H|$ and $|V|\equiv n$.
Without loss of generality we will also assume that:
$I=\{1,2,\dots,|I|\}$,  $H=\{|I|+1,|I|+2,\dots, |I|+|H|\}$
and $O=\{|I|+|H|+1, |I|+|H|+2,\dots |I|+|H|+|O|\}$.

The learning process is based on the training set $T_N=\{\,
(e^\kappa,y^\kappa)\in \mathbb R^{|I|}\times \mathbb R^{|O|}
\mid \kappa=1,\dots N\,\}$. Given an input vector $z=(z_1,z_2,\dots
z_{|I|})$, the output associated with the vertices of the graph is computed
as follows\footnote{We are using here the Iverson's notation:
 Given a statement $A$, we set $\(A)$ to $1$ if $A$ is true and to
$0$ if $A$ is false}:
\begin{equation}
  x_i(z)=z_i\(i\in I)+f_i(a_i)\(i\notin I),
  \label{comp-flow}
\end{equation}
with $a_i=\sum_{j\in\pa(i)} w_{ij} x_j +b_i$, where $\pa(i)$ are the parents
of neuron $i$, and $f_i\colon\Omega_\Lambda\to\mathbb R$
are one dimensional real functions; $\Omega_\Lambda:=[-\Lambda,\Lambda]$, with
$\Lambda$ chosen big enough, so that
Eq.~(\ref{comp-flow}) is always well defined. Now let $f=(f_1,f_2,\dots, f_n)$ and define the output function of the
network $F(\cdot, w,b;f)\colon \mathbb R^{|I|}\to \mathbb R^{|O|}$ by
\[F_i(z,w,b;f):= x_{i+|I|+|H|}(z),\quad i=1,\dots, |O|.\]
The learning problem can then be formulated as a
double optimization problem defined  on both the weights $w$, $b$
and on the activation functions $f_i$. It is worth mentioning that
while the optimization on the weights of the graph reflects all important 
issues connected with the powerful representational properties of deep nets, 
the optimal discovery of the activation functions are somewhat related
to the framework of kernel machines. 
Such an optimization is defined
with respect to the following objective function:
\begin{equation}
E(f;w,b):=\frac{1}{2}\sum_{i=1}^n (Pf_i, P f_i)
+\sum_{\kappa=1}^N V(e^\kappa, y^\kappa, F(e^\kappa, w,b;f)),
\end{equation}
which accumulates the empirical risk and a regularization term based 
regularization operators~\cite{SS-NN-1998}. Here, 
we indicate with $(\cdot,\cdot)$ the standard inner product of
$L^2(\Omega_\Lambda)$, with $P$ a differential operator of degree $p$,
while $V$ is a suitable loss function.

\begin{figure}
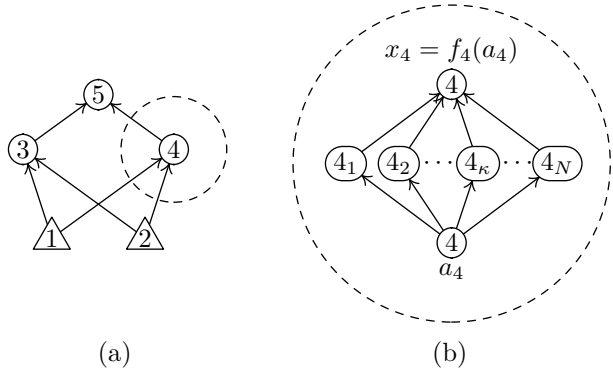


  \hbox to \hsize{\hfil \vbox{
      \halign to .5\hsize{\hfil#\hfil&\qquad\hfil#\hfil\cr
                         $\vcenter{\|1|}$&$\vcenter{\|2|}$\cr
                         \noalign{\medskip}
                         (a)&(b)\cr}}\hfil}

\caption{(a)\enspace A simple network architecture; the output
  evaluated using Eq.~(\ref{comp-flow}) is
  $x_5(z_1,z_2)=f_5(w_{53}f_3(w_{31}z_1+w_{32}z_2+b_3)+
  w_{54}f_4(w_{41}z_1+w_{42}z_2+b_4)+b_5)$. (b)\enspace Highlight of the
  structure of neuron $4$ (encircled in the dashed line)
  of (a): The activation function $f_4$
  of the neuron is computed as an expansion over the training set. Each
  neuron $4_j$, $j=1,\dots,N$ in the figure corresponds to
  the term $g(a_4-a_4^j)$ in Eq.~(\ref{represent-theo}).}
\label{node-fig}
\vskip -0.2in
\end{figure}

Clearly, one can optimize $E$ by independently checking the stationarity 
with respect to the weights associated with the neural connections 
and the stationarity with respect to the activation functions.
Now we show that the stationarity condition of $E$ with respect
to the functional variables $f$ (chosen in a functional space $X_p$ that
depends on the order of differential operator $P$)
yields a solution that is very related to classic case of kernel machines
that is addressed in~\cite{SS-NN-1998}. If we consider 
a variation $v_i\in C^\infty_c(\Omega_\Lambda)$ with vanishing
derivatives on the boundary 
\footnote{Here, we are assuming here that
the values of the functions in $X_p$ at the boundaries together with
the derivatives up to order $p-1$ are fixed.}
of $\Omega_\Lambda$ up to order $p-1$
and define $\varphi_i(t):=E(f_1,\dots,f_i+tv_i,\dots, f_n; w,b)$. The first 
variation of the functional $E$ along $v_i$ is therefore $\varphi_i'(0)$.
When using arguments already discussed in related papers 
~\cite{Poggio_ieee90,PoggioGirosiNC95,SS-NN-1998}
we can easily see that
\[\varphi_i'(0)=\int_{\Omega_\Lambda}
  \Big(L f_i(a)+\sum_{\kappa=1}^N\alpha^\kappa_i
  \delta_{a^\kappa_i}(a)\Big)v_i(a)\, da,\]
where $\alpha^\kappa_i=\nabla_F V\cdot \partial_{f_i} F$ and $L=P^*P$,
$P^*$ being the adjoint operator of $P$.
We notice in passing  that the functional dependence of $E$ on $f$ 
is quite involved, since it depends on the compositions of
liner combinations of the functions $f_i$ (see Figure~\ref{node-fig}--(a)).
Hence, the given expression of the coefficients $\alpha^\kappa_i$ is a
rather a formal equation that, however, dictates the structure of the solution.
\def\figdir{.}
\def\|#1|{\hbox{ \includegraphics{\figdir/fun-#1.mps}}}

The stationarity conditions $\varphi_i'(0)=0$ reduce to the
following Euler-Lagrange (E-L) equations
\begin{equation}
  L f_i(a)+\sum_{\kappa=1}^N\alpha^\kappa_i
  \delta_{a^\kappa_i}(a)=0,\quad i=1\dots n ,
 \label{ELE}
\end{equation}
where $a^\kappa_i$ is the value of the activation function
on the $\kappa$-th example of the training set.
Let $g$ be the Green function of the operator
$L$, and let be $k$ the solution of $Lk=0$.  Then, we can promptly see that
\begin{equation}
  f_i(a)=k(a)-\sum_{\kappa=1}^N \alpha^\kappa_i
  g(a-a^\kappa_i)
  \label{represent-theo}
\end{equation}
is the general form of the solution of Eq.~(\ref{ELE}). Whenever
$L$ has null kernel, then this solution is reduced to an expansion of the Green function
over the points of the training set. For example, this happens in the case of
the pseudo differential operator that originates the Gaussian as the Green function. 
If we choose $P=d/dx$, then $L=-d^2/dx^2$.
Interestingly, the Green function of the second derivative is the rectifier 
$g(x)=-\frac{1}{2}(|x|+x)$ and, moreover, we have $k(x)=mx+q$. 
In this case
\begin{equation}
  f_i(a)=\theta_i a+\nu_i+\frac{1}{2}
  \sum_{\kappa=1}^N \alpha^\kappa_i|a-a^\kappa_i|,
\end{equation}
where $\theta_i=m+\frac{1}{2}\sum_{\kappa=1}^N \alpha_i^\kappa$, while  $\nu_i=
q-\frac{1}{2}\sum_{\kappa=1}^N\alpha^\kappa_i a^\kappa_i$.
%
Because of the representation structure expressed by Eq.~(\ref{represent-theo}),
the objective function the original optimization problem collapses to a standard 
finite-dimensional optimization on\footnote{Here we
 omit the dependencies of the optimization function from the parameters that
defines $k$.}
\[\hat E(\alpha,w,b):=
    E\Big(k(a)-\sum_\kappa \alpha^\kappa g(a-a^\kappa);w,b\Big)
    =R(\alpha)+\sum_{\kappa=1}^N V(e^\kappa, y^\kappa,
    \hat F(e^\kappa, w,b;\alpha));\]
here $R(\alpha)$ is the regularization term and
$\hat F(e^\kappa, w,b;\alpha):=F\big(e^\kappa,w,b;k(a)-\sum_\kappa
\alpha^\kappa_i g(a-a^\kappa_i)\big)$.
This collapse of dimensionality is the same which leads to the dramatic 
simplification that gives rise to the theory of kernel machines. Basically, 
in all cases in which the Green function can be interpreted as a kernel,
this analysis suggests the neural architecture depicted in 
Figure~\ref{node-fig}, where we can see the integration of graphical
structures, typical of deep nets, with the representation in the
dual space that typical of kernel methods.

We can promptly see that the idea behind kernel-based deep networks can 
be extended to cyclic graphs, that is to recurrent neural networks. In that case,
the analogous of Eq.~(\ref{comp-flow}) is:
\[h_i^{t+1}=f_i(a_i^{t+1});\qquad
 a_i^{t+1}=b_i+\sum_{j\in\pa_{t\to t+1}(i)}w_{ij} h^t_j
    +\sum_{j\in\pa_{t+1}(i)}u_{ij} x^{t+1}_j.\]
Here we denote with $x_i^t$ the input at step $t$ and with $h_i^t$ the
state of the network.
The set $\pa_{t\to t+1}(i)$ contains the vertices $j$ that are parents
of neuron $i$; the corresponding  arcs $(j,i)$ are associated with a delay,
while $\pa_t(i)$ vertices $j$ with non-delayed arcs $(j,i)$.
The extension of learning in KBDNN to the case of recurrent nets is a straightforward
consequence of classic Backpropagation Through Time. 


\section{Approximation and algorithmic issues}
\label{approximation}
The actual experimentation of the model described in the
previous section  requires to deal with a number of important
algorithmic issues. In particular, we need to address the 
typical problem associated with the kernel expansion over the
entire training set, that is very expensive in computational terms. 
However, we can early realize that KBDNNs only require to 
express kernel in 1-D, which dramatically simplify the kernel 
approximation. Hence, instead of expanding $f_i$ over the entire
training set, we can use a number of points $d$ with $d \ll N$.
This means that the expansion in Eq.~(\ref{represent-theo})
is approximated as follows
\begin{equation}\label{approx-expansion}
  f_i(a)\approx k(a)-\sum_{k=1}^d \chi_i^k g(a-c_i^k),
\end{equation}
where $c_i^k$ and  $\chi_i^k$ are the centers and parameters of the expansion, respectively. Notice that $\chi_i^k$ are replacing
$\alpha^\kappa_i$ in the formulation given in Section~\ref{theory}). We consider $c_i^k$ and $\chi_i^k$ as parameters to be learned,
and integrate them in the whole optimization scheme.

In the experiments described below we use the rectifier (ReLU) as Green
function ($g(x)=-\frac{1}{2}(|x|+x)$) and neglect the linear terms from
both $g(x)$ and $k(x)$. We can easily see that this is compatible with
typical requirements in machine learning experiments, where
in many cases the expected solution is not meaningful with 
very large inputs. For instance, the same assumption is typically at the
basis of kernel machines, where the asymptotic behavior is not typically
important. The regularization term $R(\chi)$ can be inherited from the
regularization operator $P$. For the experiments carried out in 
this paper we decided to choose the $\ell_1$ 
norm\footnote{This choice is due to the fact that we want 
to enforce the sparseness of $\chi$, i.e. to use the smallest number of terms in
  expansion~\ref{approx-expansion}.}:
\[R(\chi)\approx \lambda_\chi\sum_{\scriptstyle1\le k\le d \atop\scriptstyle1\le i\le n}
  |\chi_i^k|,\]
with $\lambda_\chi\in \mathbb R$ being an hyper-parameter that measures the
strength of the regularization.


%
%

In a deep architecture, when stacking multiple layers of kernel-based units,
the non-monotonicity of the activation functions implies the absence of guarantees 
about the interval on which these functions operate, thus requiring them to be responsive
to very heterogeneous inputs.
In order to face this problem and to allow kernel-based units to concentrate their representational power on limited input ranges, it is possible to apply a normalization \cite{ioffe2015batch} to the input of the function.
In particular, given $f_i(a_i^\kappa)$,  $a_i^\kappa$ can be normalized as:
\[\hat a_i^\kappa = \gamma_i\frac{(a_i^\kappa-\mu_i)}{\sigma_i}+\beta_i;
  \qquad \hbox{where}\qquad
\mu_i  = \frac{1}{N}\sum_{\kappa=i}^N a_i^\kappa,\qquad 
\sigma_i = \frac{1}{N} \sum_{\kappa=i}^N ( a_i^\kappa - \mu_i)^2;\] 

%
%

while $\gamma_i$ and $\beta_i$ are additional trainable parameters.

\section{Experiments}
\label{expriments}

\begin{figure}                                                                  
             
        \begin{subfigure}{0.43\linewidth}
			\includegraphics[width=0.98\linewidth]{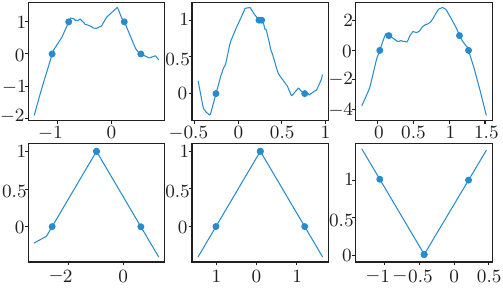}
			\caption{}
        	\label{fig:xorsfig2}
        \end{subfigure}%
        \begin{subfigure}{0.5\linewidth}
       		\includegraphics[width=0.98\linewidth]{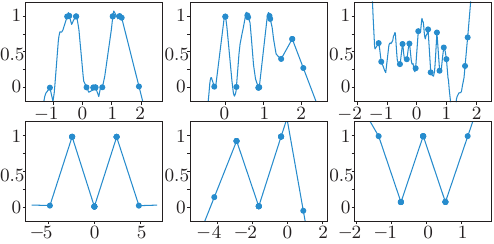}
			\caption{}
			 \label{fig:xorsfig4}
        \end{subfigure}
          \caption{\textbf{XOR.\enspace} The plots show the activation         
        	functions learned by the simplest KBDN which                          
        	consists of one unit only for the 2-dim (\ref{fig:xorsfig2}) and 4-dim (\ref{fig:xorsfig4}) XOR. The first/second row refer to               
        	experiments with without/with regularization,                          
        	whereas the three columns correspond with the chosen number of                   
        	point for the expansion of the Green function                          
        	$d= 50, 100, 300$.}                                                    
        
        \label{fig:xor2-4}   
\end{figure}
      
We carried out  several experiments in different learning settings to
investigate the effectiveness of the KBDNN with
emphasis on the adoption of kernel-based units in recurrent
networks for capturing long-term dependences. 
Clearly, KBDNN architectures require to choose both the
graph and the activation function. As it will be clear in the 
reminder of this section, the interplay of these choices
leads to gain remarkable properties.
%
\begin{figure}
	    \centering
		\includegraphics[width=.5\linewidth]{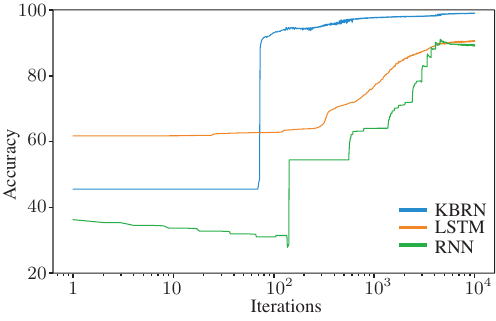}
		\caption{\textbf{Charging Problem.}\enspace 
				The plot shows the accuracy
				obtained a by recurrent nets which classic sigmoidal unit,
				LSTM cell, and KB unit.  
				The horizontal axis is in logarithmic scale.
				}
	\label{fig:cp}
\end{figure}


\medskip
\noindent
\textbf{The XOR problem.\enspace}
We begin presenting a typical experimental set up in the 
classic XOR benchmark. 
In this experiment we chose a single unit with
the Green function $g(z) = |z|$, so as 
$y=f(w_1z_1 + w_2z_2 +b)$ turns out to be
\begin{align*}
	y = 
	\sum_{k=1}^{d} 
	\chi^k |w_1 z_1+w_2 z_2 + b - c^k|
\end{align*}
where $w_1$,$w_2$ and $b$ are trainable variables and the 
learning of $f$ corresponds with the discovery of 
both the centroids $c^k$ and the associated weights $\chi^k$. 
The simplicity of this learning task allows us to underline some interesting properties of KBDNNs. We carried out experiment
by selecting a number of points for the expansion of the 
Green function that ranges from $50$ to $300$. This was
done purposely  to assess the regularization capabilities of the
model, that is very much related to what typically happens with
kernel machines. In Figure~\ref{fig:xor2-4}, we can see the neuron
function $f$ at  the end of the learning process under  different settings. In the different columns, we plot function $f$ 
with a different numbers $d$ of clusters, while the two rows
refer to experiments carried out with and without regularization.
As one could expect, the learned activation functions become more
and more complex as the number of clusters increases. However,
when performing regularization, the effect of the kernel-based
component of the architecture plays a crucial role by smoothing
the functions significantly. 
%
%

\medskip
\noindent
\textbf{The charging problem.\enspace}
Let us consider a dynamical system which generates a
Boolean sequence according to the model
\begin{equation}
	\begin{split}
	h_t &= x_{t} + \(h_{t-1} - 1 > 0)\cdot(h_{t-1}-1)\\
	y_t &= \(h_t > 0),
	\end{split}
	\label{charge-eq}
\end{equation}
where $h_{-1}=0$, $x = \langle x_t\rangle$ is a sequence of integers
and $y = \langle y_t\rangle$ is a Boolean sequence, that is
$y_t \in \{0,1\}$. An example of sequences generated by this system is
the following:
\[\vbox{\halign{ $\hfil#=$ &&\ \hfil$#$\hfil\cr
 t&0&1&2&3&4&5&6&7&8&9&10&\dots\cr
 x_t&0&0&0&4&0&0&0&0&0&0&0&\dots\cr
 y_t&0&0&0&1&1&1&1&0&0&0&0&\dots\cr}}.\]
Notice that the system keeps memory when other
$1$ bit are coming, that is
\[\vbox{\halign{ $\hfil#=$ &&\ \hfil$#$\hfil\cr
 t&0&1&2&3&4&5&6&7&8&9&10&\dots\cr
 x_t&0&0&0&4&0&2&0&0&0&0&0&\dots\cr
 y_t&0&0&0&1&1&1&1&1&1&0&0&\dots\cr}}\]
The purpose of this experiment was that of checking
what are the learning capabilities of KBRN to approximate
sequences generated according to Eq.~\ref{charge-eq}.
The intuition is that a single KB-neuron is capable to \textit{charge} 
the state according to an input, and then to \textit{discharge} it 
until the state is reset.
We generated sequences $\langle x_{t} \rangle$ of length $L = 30$. 
Three random element of each sequence were set with a random number ranging from $0$  to $9$. We compared KBRN, RNN 
with sigmoidal units, and recurrent with LSTM cells, 
with a single hidden unit. We used a KBRN unit with $d=20$ centers to approximate the activation function. The algorithm used for optimization used the Adam algorithm with $\lambda = 0.001$ in all cases. Each model was  trained for $10000$ iterations with mini-batches of size $500$. Figure~\ref{fig:cp} shows the accuracy on a randomly generated test set of size $25000$ during the training
process. The horizontal axis is in logarithmic scale. The horizontal axis is in logarithmic scale. 

\begin{figure}
	\centering
	\includegraphics[width=0.7\hsize]{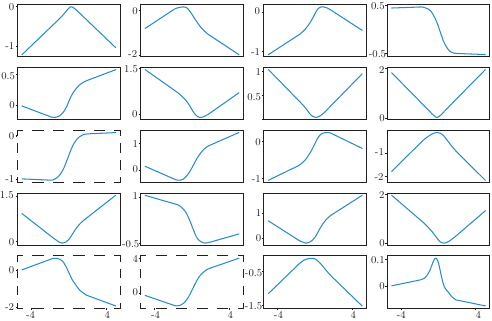}
	\caption{\textbf{Activation functions.\enspace} The $20$ activation
          functions corresponding with the problem of capturing long-term dependencies
          in sequences that are only discriminated by the first two bit ($\equiv$ function).
          All functions are plotted in the interval $[-4,4]$. The functions
          with a dashed frame are the ones for which $|f'|>1$ in some
          subset of $[-4,4]$.}
	\label{fig:activation:ltm}
      \end{figure}

\begin{figure*}
	\centering
	\includegraphics[width=\textwidth]{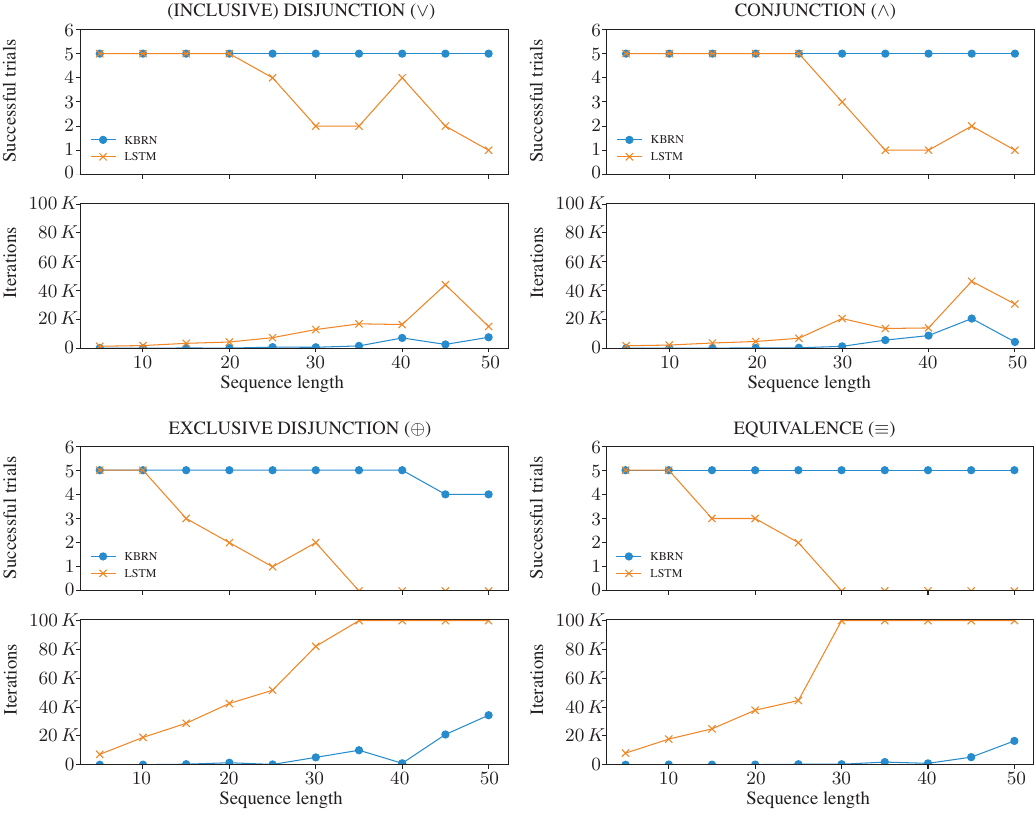}
	\caption{\textbf{Capturing  Long-Term dependencies.\enspace} Number of successful trials and average number of iterations for a classification problem when the $\lor$, $\land$, $\oplus$ and $\equiv$ functions are used to determine the target, given the first two discriminant bits.}
	\label{fig:ltm_bool_fun}
\end{figure*}

\medskip\noindent
\textbf{Learning Long-Term dependencies.\enspace}
We carried out a number of experiments aimed at investigating the
capabilities of KBRN in learning tasks where we need to capture 
long-term dependencies. The difficulties of solving similar problems
was addressed in~\cite{Bengio_trnn93} by discussions on 
gradient vanishing that is mostly due to the monotonicity of the
activation functions. The authors also provided very effective
yet simple benchmarks to claim that classic recurrent networks 
are unable to classify sequences where the distinguishing information
is located only at the very beginning of the sequence; the rest of the
sequence was supposed to be randomly generated.
We defined a number of  benchmarks inspired
by the one given in~\cite{Bengio_trnn93}, where the decision 
on the classification of sequence $\langle x_{t} \rangle$ is contained
in the first $L$ bits of a Boolean sequence of length $T \gg L$.
%
We compared KBRN and recurrent nets with LSTM cells using 
an architecture where both networks were based on $20$ hidden units. 
We used the Adam algorithm with $\lambda = 0.001$ in all cases. Each model was trained for a maximum of $100,000$ iterations with mini-batches of size $500$; for each iteration, a single weight update was performed. For the LSTM cells, we used the standard implementation provided by TensorFlow (following 
\cite{zaremba2014recurrent}). 
For KBRN we used a number of centroids $d=100$ and the described
normalization.

We generated automatically a set of benchmarks with $L=2$ and 
variable length $T$, where the binary sequences $\langle x_t \rangle$ 
can be distinguished when looking simply at the first two bits, while the 
the rest is a noisy string with uniformly random distribution. 
Here we report some of our experiments when choosing the 
first two discriminant bits according to the
$\lor$, $\land$, $\oplus$ and $\equiv$ functions. 

\begin{figure}
	\centering
	\includegraphics[width=.5\hsize]{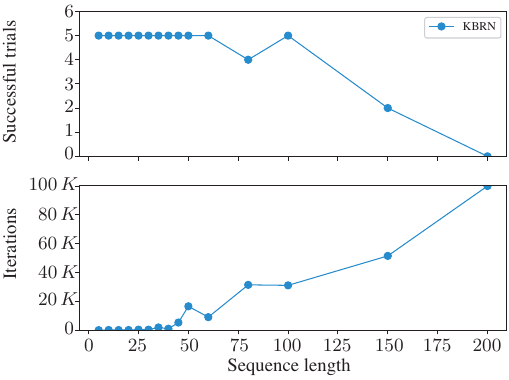}
	\caption{\textbf{Capturing  Long-Term dependencies.\enspace} Number of successful trials and average number of iterations when facing the $\equiv$ problem with sequences of length ranging from 5 to 200, when the distinguishing information is located in the first two bits. }
	\label{fig:lt_iff_longshot}
\end{figure}

For each Boolean function, that was supposed to be learned, and for several sequence lengths (up to 50), we performed 5 different runs, with different initialization seeds.  
A trial was considered successful if the model was 
capable of learning the function before the maximum allowed number of iterations was reached. In Figure~\ref{fig:ltm_bool_fun} we present the results of these experiments. Each of the four quadrants of Figure~\ref{fig:ltm_bool_fun} is relative to a different Boolean function, and reports two different plots. The first one has the sequence length on the $x$-axis and the number of successful trials on the $y$-axis. The second plot has the sequence length on the $x$-axis and, on the $y$-axis, the average number of iterations required  to solve the task.  The analysis of these plots allows us to draw
ta couple of interesting conclusions: \textit{(i)} KBRN architectures are capable of solving the problems in almost all cases, regardless of the sequence length, while recurrent networks with LSTM cells
started to experiment difficulties for sequences longer than 30, and \textit{(ii)}, 
whenever convergence is achieve,  KBRN architectures converge significantly faster than 
LSTM. 
In order to investigate with more details the capabilities of KBRN of handling very long sequences, we carried out another experiment, that was based on the benchmark that KBRN solved with more difficulty, namely the equivalence ($\equiv$) problem. We carried out a processing over 
 sequences with length $60, 80, 100, 150$, and $200$. In Figure~\ref{fig:lt_iff_longshot}, we report the results of this experiment. As we can see, KBRN are capable of solving the task even with sequences of length 150, eventually failing with sequences of length 200.


\section{Conclusions}
\label{conclusion}
In this paper we have introduced Kernel-Based Deep Neural Networks.
The proposed KBDNN model is  characterized
by the classic primal representation of deep nets, that is enriched
with the expressiveness of activation functions given by kernel
expansion. The idea of learning the activation 
function is not entirely new.  
However, in this paper we have shown that
the KBDNN representation turns out to be the solution of a 
general optimization problem, in which both the weights, 
that belong to a finite-dimensional space, and the activation function, 
that are chosen from a functional space
are jointly determined. This bridges naturally the powerful 
representation capabilities of deep nets with the elegant and effective
setting of kernel machines for the learning of the neuron functions. 

A massive experimentation of KBDNN is still required to assess the
actual impact of the appropriate activation function in real-world problems.
However, this paper already proposes a first  important conclusion 
which involves recurrent networks, that are based on this kind of 
activation function. In particular, we have provided both theoretical
and experimental evidence to claim that the KBRN architecture
exhibits an ideal computational structure to deal with classic
problems of capturing long-term dependencies.


\bibliographystyle{plain}
\bibliography{nn,References}

\end{document}